\documentclass[11pt]{article}

\usepackage[margin=1in]{geometry}
\usepackage[T1]{fontenc}
\usepackage[utf8]{inputenc}
\usepackage{newtxtext,newtxmath}
\usepackage{amsmath}
\usepackage{booktabs}
\usepackage{graphicx}
\usepackage{microtype}
\usepackage[hidelinks]{hyperref}

\newcommand{\dist}[1]{\textbf{#1}}

\title{Kathleen Writes: Autoregressive Generation and Data Scaling\\
Without Attention}
\author{George Fountzoulas\\
\small Department of Computer Engineering \& Informatics\\
\small Frederick University, Nicosia, Cyprus\\
\small \texttt{george.fountzoulas.research@gmail.com}}
\date{August 2026}

\begin{document}
\maketitle

\begin{abstract}
Papers 1--2 of the Kathleen series showed that a byte-level,
attention-free architecture built from a wavetable encoder and
multi-scale reverberant state can match strong baselines on
classification at ${\sim}$450--700K parameters, without pretraining.
The open question was whether the same ingredients can
\emph{generate}. We answer in three parts.
\textbf{(1) Scaling.} On byte-level language modeling (WikiText-103,
raw UTF-8, no tokenizer), the reverberant model beats a
parameter-matched transformer at every dataset scale we measured ---
2, 8, 32, 128 and 512\,MB --- e.g.\ 1.84 vs 2.04 bits/byte at 512\,MB
with ${\sim}$0.5M parameters, while training within a free-tier GPU
budget. Put as data efficiency: the transformer needs more than
512\,MB to match what the attention-free model learns from 32\,MB.
The attention baseline shows the steeper log-log slope ($-0.107$ vs
$-0.074$) --- the shape of catching up from behind --- but both curves
flatten toward the same fixed-capacity ceiling, and its rate of
catching up itself decays; within the on-device regime the
attention-free model is ahead everywhere.
\textbf{(2) Measurement.} Perplexity does not measure whether
generated text \emph{reads} like text. We introduce FORM DISTANCE, a
non-parametric, gaming-resistant instrument: nine statistical axes of
human text (lexical richness, flow, local repetition, bigram/trigram
plausibility, \ldots) define a reference cloud; a generator is scored
by its distance to that cloud, normalized so human text sits at 1.0.
Five constructed fakes (word soup, parrot, drone, shuffle, trigram
babble) are all rejected by the instrument.
\textbf{(3) Generation.} On the frozen instrument, decoding policy
dominates architecture: widening the sampler (top-1000, $T=1.15$)
halves the same model's distance, $3.17 \to 1.52$ --- an earlier
eight-axis version of the instrument read this policy as fully human
(1.00), and the disagreement with a human reader exposed the gaming
channel (invented trigrams) that the ninth axis now closes. A
retrieval-augmented decoding scheme --- corpus phrases proposed during
generation and interleaved with free words --- takes the frozen model
further, $1.52 \to 1.14$, with confidence intervals separated and no
training step involved; the ablation attributes the gain to the
sparse phrase dose itself, not to the hidden-state gate that selects
among phrases. The gain has a sharp boundary condition: the phrases
must come from the model's \emph{own} training corpus --- a
$40\times$ larger foreign library helps not at all, an effect the
attention twin shares (we audit both the instrument and the
architecture), consistent with in-context integration being a
capability of scale that 3M-parameter models of either family lack.
We also report, honestly, four architectural additions that did
\emph{not} help at this scale, and the negative result that a
computed (gradient-free) lexicon reaches 94\% of a learned embedding
table's top-1 accuracy at one fifth of the parameters, but not its
perplexity. Everything runs offline on commodity hardware; all
experiments are reproducible on a free Kaggle T4.
\end{abstract}

\section{Introduction}

The first two papers of this series established that a small,
attention-free, tokenizer-free architecture can \emph{read}:
operating directly on raw UTF-8 bytes, a trunk built from a wavetable
byte encoder and a multi-scale reverberant state matched or beat much
larger tokenized baselines on sentiment and topic classification at
${\sim}$450--700K parameters --- and, in the second paper
\cite{kathleen2}, did so with gradient-based pretraining eliminated
entirely. The encoder derives all 256 byte vectors from a single
learnable vector by DFT phase rotation; the reverberant state
replaces attention with content-gated exponential decay at three time
constants --- a fading memory whose forgetting rate is decided by the
content itself. What those papers deliberately left open is the
harder verb: can the same ingredients \emph{write}?

Answering that question turns out to require answering three at once.
First, \textbf{scaling}: a generator lives or dies by its language
model, so we must know whether the attention-free trunk models
next-byte prediction competitively, and --- since no small model is
trained once --- how that competitiveness moves with data. Second,
\textbf{measurement}: perplexity measures prediction, not whether
sampled text \emph{reads} like text, and the fashionable
alternative --- asking a large language model to judge --- is circular
for a project whose premise is independence from large models, and
unusable offline. We need an instrument that is cheap, frozen, and
hard to game. Third, \textbf{attribution}: when generation improves,
which knob did the work --- the architecture, the decoding policy, or
memory bolted on at decode time? Small models are where this question
can actually be answered, because every knob can be swept
exhaustively on free-tier hardware.

The design constraints of the series carry over unchanged:
${\sim}$0.5M parameters at byte level and $\le{\sim}$3M at word
level; everything offline; every experiment reproducible on a free
Kaggle T4; no external teacher model anywhere in the loop. Within
those constraints this paper contributes: (a) a five-point
data-scaling study, 2--512\,MB, where the attention-free trunk beats
a parameter-matched transformer at every point, replicated across
seeds and a second dataset, and corroborated by a frozen-probe
transfer study; (b) FORM DISTANCE, a non-parametric, validated,
gaming-resistant instrument for ``reads like text'', including the
honest record of the one time it was gamed and how it was patched;
and (c) a decode-time attribution study whose headline is
uncomfortable and useful: at this scale, form is won by the decoding
policy and by retrieved phrases, not by architectural additions to
the trunk --- we report four pre-registered additions that did not
help.

\section{Architecture (shared ingredients)}

\begin{itemize}
\item \textbf{ByteRotate encoder} --- one learnable vector
  $w \in \mathbb{R}^d$; byte $b$ gets
  $\mathrm{irfft}(\mathrm{rfft}(w) \cdot e^{2\pi i \, b f / 256})$.
  128 parameters cover the whole byte alphabet. (Carried from
  Paper 1.)
\item \textbf{Multi-scale reverb} --- three banks with
  content-dependent decay $\gamma \in [0.50,0.90]$ / $[0.90,0.99]$ /
  $[0.95,0.9995]$; recurrence
  $s_t = \gamma_t s_{t-1} + (1-\gamma_t)\,v_t$. This is the attention
  substitute: a fading, content-gated memory at three time scales.
\item \textbf{Reverb block} --- LayerNorm $\to$ reverb $\parallel$
  causal depthwise conv $\to$ learned gate mixing the two $\to$ FFN.
  Three blocks, $d=128$.
\item \textbf{Fast scan} (new, engineering): the reverb recurrence
  evaluated in chunks of 16 via a cumprod/cumsum closed form;
  equivalence to the sequential loop verified to $10^{-7}$;
  ${\sim}$10--50$\times$ wall-clock speedup. This is what makes the
  512\,MB point trainable on a free T4. (Appendix~\ref{app:scan}.)
\item Byte-level LM: 565{,}888 params (reverb) vs 463{,}360
  (attention baseline: same skeleton with 4-head causal SDPA in place
  of reverb+conv). Word-level variant for the generation studies:
  3.08M params, 10K vocab, WikiText-2.
\end{itemize}

\section{Scaling without attention (the road)}
\label{sec:scaling}

\paragraph{Protocol.} WikiText-103 \cite{merity2017wikitext} as raw
UTF-8 bytes. Training slices of 2, 8, 32, 128, 512\,MB; fixed test
slice (1\,MB); 2 epochs, seq 256 bytes, AdamW, cosine schedule,
identical recipe both arms; pre-registered verdict thresholds before
running. Free-tier hardware (T4/H100 mix; bits-per-byte is
hardware-independent).

\begin{table}[ht]
\centering
\caption{Byte-level scaling road, seed 42 (bits per byte, lower is
better).}
\label{tab:road}
\begin{tabular}{rccc}
\toprule
MB & REVERB bpb & ATTN bpb & gap \\
\midrule
2   & 2.844 & 3.586 & 0.74 \\
8   & 2.239 & 3.081 & 0.84 \\
32  & 2.010 & 2.416 & 0.41 \\
128 & 1.904 & 2.155 & 0.25 \\
512 & 1.842 & 2.040 & 0.20 \\
\bottomrule
\end{tabular}
\end{table}

Log-log slopes (bpb vs MB): REVERB $-0.074$, ATTN $-0.107$.

\begin{figure}[ht]
\centering
\includegraphics[width=0.85\linewidth]{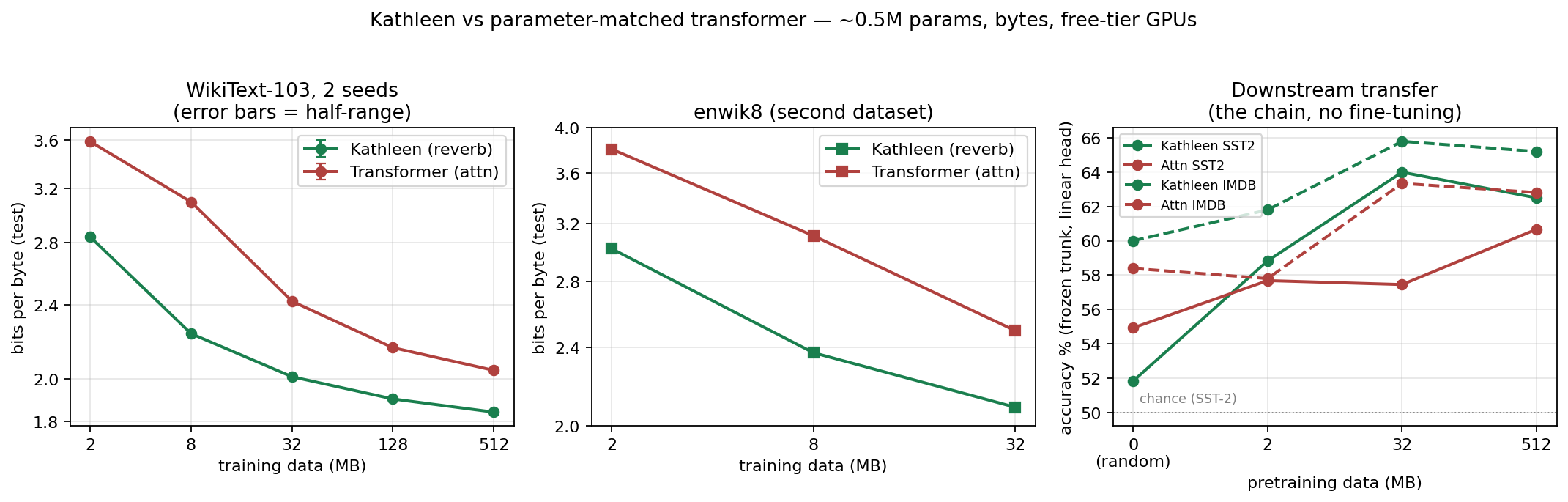}
\caption{The road: bits per byte vs training data (log-log), both
arms, five points, with two-seed error bars.}
\label{fig:road}
\end{figure}

\paragraph{Reading.} (i) The attention-free model wins every point
across three orders of magnitude of data. (ii) Stated as data
efficiency: the transformer trained on 512\,MB (2.043 bpb) does not
reach the reverb model trained on 32\,MB (2.011 bpb) --- a
$\ge 16\times$ data advantage at equal parameters. (iii) The
attention arm shows the steeper log-log slope ($-0.107$ vs $-0.074$),
but this must not be over-read: a steeper slope is what catching up
from behind looks like, and both curves flatten together --- the
signature of the shared ${\sim}$0.5M-parameter capacity ceiling, not
an architectural verdict. Extrapolating the \emph{final} interval
puts any crossover in the hundreds-of-GB regime, far outside the
on-device envelope this work targets. (iv) What this experiment does
\textbf{not} measure, by design: large-scale behavior when parameters
grow with data (the compute-optimal frontier
\cite{kaplan2020,hoffmann2022}). No conclusion about that regime ---
in either direction --- follows from these curves; related
recurrences (the SSM family \cite{gu2022s4,gu2023mamba,peng2023rwkv})
are known to track transformers to billions of parameters, and the
corresponding experiment for this architecture is future work, not a
concession.

\paragraph{Robustness (measured 2026-08-02, Kaggle T4, single
session).} \emph{Second seed.} The full road repeated with seed 43
reproduces seed 42 almost exactly (e.g.\ 128\,MB REVERB: 1.9036 vs
1.9039). Two-seed means:

\begin{table}[ht]
\centering
\caption{Two-seed means $\pm$ half-range.}
\label{tab:seeds}
\begin{tabular}{rcc}
\toprule
MB & REVERB bpb & ATTN bpb \\
\midrule
2   & $2.8365 \pm 0.0076$ & $3.5884 \pm 0.0025$ \\
8   & $2.2359 \pm 0.0030$ & $3.0922 \pm 0.0116$ \\
32  & $2.0109 \pm 0.0013$ & $2.4217 \pm 0.0057$ \\
128 & $1.9038 \pm 0.0001$ & $2.1600 \pm 0.0052$ \\
512 & $1.8430 \pm 0.0016$ & $2.0427 \pm 0.0023$ \\
\bottomrule
\end{tabular}
\end{table}

Seed-level slopes agree to the third decimal (REVERB
$-0.0744/-0.0732$, ATTN $-0.1071/-0.1072$): the curve is a property
of the system, not of the seed.

\emph{Second dataset.} enwik8 (Wikipedia XML, unclean bytes),
2/8/32\,MB:

\begin{table}[ht]
\centering
\caption{enwik8 robustness check.}
\label{tab:enwik8}
\begin{tabular}{rccc}
\toprule
MB & REVERB bpb & ATTN bpb & gap \\
\midrule
2  & 3.020 & 3.806 & 0.79 \\
8  & 2.372 & 3.112 & 0.74 \\
32 & 2.090 & 2.497 & 0.41 \\
\bottomrule
\end{tabular}
\end{table}

Slopes on enwik8: REVERB $-0.133$, ATTN $-0.152$ --- near-parallel;
the ``attention is steeper'' reading is itself dataset-dependent.
Every completed point across both datasets and both seeds: REVERB
ahead.

\paragraph{Downstream transfer (BRIDGE, H100, frozen-probe
protocol).} Trunks pretrained at 0/2/32/512\,MB, frozen; a
\emph{linear head only} (258 params) trained on SST-2
\cite{socher2013sst} and IMDB \cite{maas2011imdb}:

\begin{table}[ht]
\centering
\caption{Frozen-probe transfer (accuracy, \%).}
\label{tab:bridge}
\begin{tabular}{rcccc}
\toprule
pretrain MB & SST-2 REV & SST-2 ATT & IMDB REV & IMDB ATT \\
\midrule
0 (random) & 51.8 & 54.9 & 60.0 & 58.4 \\
2   & 58.8 & 57.7 & 61.8 & 57.8 \\
32  & \textbf{64.0} & 57.5 & \textbf{65.8} & 63.3 \\
512 & 62.5 & 60.7 & 65.2 & 62.8 \\
\bottomrule
\end{tabular}
\end{table}

The BERT chain \cite{devlin2019bert} holds without attention:
byte-level LM pretraining alone injects sentiment into the frozen
representation ($+12$ points over the random trunk on SST-2), and the
reverb trunk transfers better than the attention twin at every
pretrained scale. The 32$\to$512\,MB plateau matches the bpb curve
--- a second, independent instrument reading the same capacity
ceiling.

\paragraph{Probe vs full recipe (the two ends of the same chain).}
These are frozen-probe numbers by design --- they isolate what
pretraining alone puts into the trunk. The other end is measured too:
the full Kathleen recipe (Arm-H: V9 byte pipeline + counted
holographic meaning fused at the classifier + a 5-stage curriculum of
hand-built linguistic lessons --- negation flips, contrast,
``despite'', concessive constructions --- + 80K Stanford phrases with
strict dev exclusion) reaches \textbf{96.4\% on SST-2 dev at 468K
parameters} (best observed run 96.56\%), with the residual errors
concentrated exactly where syntax is hardest (negation 4.5\%,
contrast 3.4\% error rates). Mechanism (this section) and ceiling
(Arm-H) bracket the claim: reading more bytes buys understanding, and
targeted lessons plus counted meaning buy the rest --- no attention
and no gradient-based pretraining anywhere in the chain.
Consistently, our earlier document-level study found IMDB plateaus
near 88\% under a frozen sentence-trained trunk regardless of
aggregation strategy (mean/top-k/MIL/subjectivity gating) --- the pp
that are missing live in task-aligned pretraining, which is precisely
the lever this section measures.

\section{FORM DISTANCE --- a meter for ``reads like text''}
\label{sec:meter}

Bits-per-byte measures how well a model predicts text; it says
nothing about whether the text the model \emph{samples} reads like
text. The two can and do come apart: every decoding-policy result in
Section~\ref{sec:generation} changes form dramatically while the
weights --- and hence the perplexity --- stay fixed. The fashionable
fix, asking a large language model to judge, is doubly unavailable
here: it is circular for a project whose premise is independence from
large models, and it does not run offline. We therefore built a
frozen instrument from surface statistics alone. (For parametric
judge-free alternatives see MAUVE \cite{pillutla2021mauve};
Section~\ref{sec:related}.)

\paragraph{Construction (v6.2).} Nine axes are computed from a
passage's surface statistics: content-word richness (RICH), trigram
diversity (FLOW), theme persistence between opening and closing
thirds (HOLD), chunk-to-chunk semantic continuity (GRAIN),
late-novelty rate (NEW), local bigram well-formedness (LOCAL),
semantic development with distance (ARC), unseen-bigram rate (CLUNK)
and unseen-trigram rate (CLUNK3). Four hundred held-out human
WikiText passages define a reference cloud in this nine-dimensional
space; a generator is scored by a Mahalanobis-style distance to the
cloud, normalized so that held-out human text reads 1.0. A companion
percentile view (typicality: human ${\approx}$ 50) gives resolution
near the cloud where distance saturates. Crucially, no axis direction
is assumed good: an axis earns its place only if both degenerate
controls fall far outside the human band on it --- word soup fails in
one direction and a parrot in the other, so ``more'' of any axis is
never the target, \emph{typicality} is.

\begin{figure}[ht]
\centering
\includegraphics[width=0.9\linewidth]{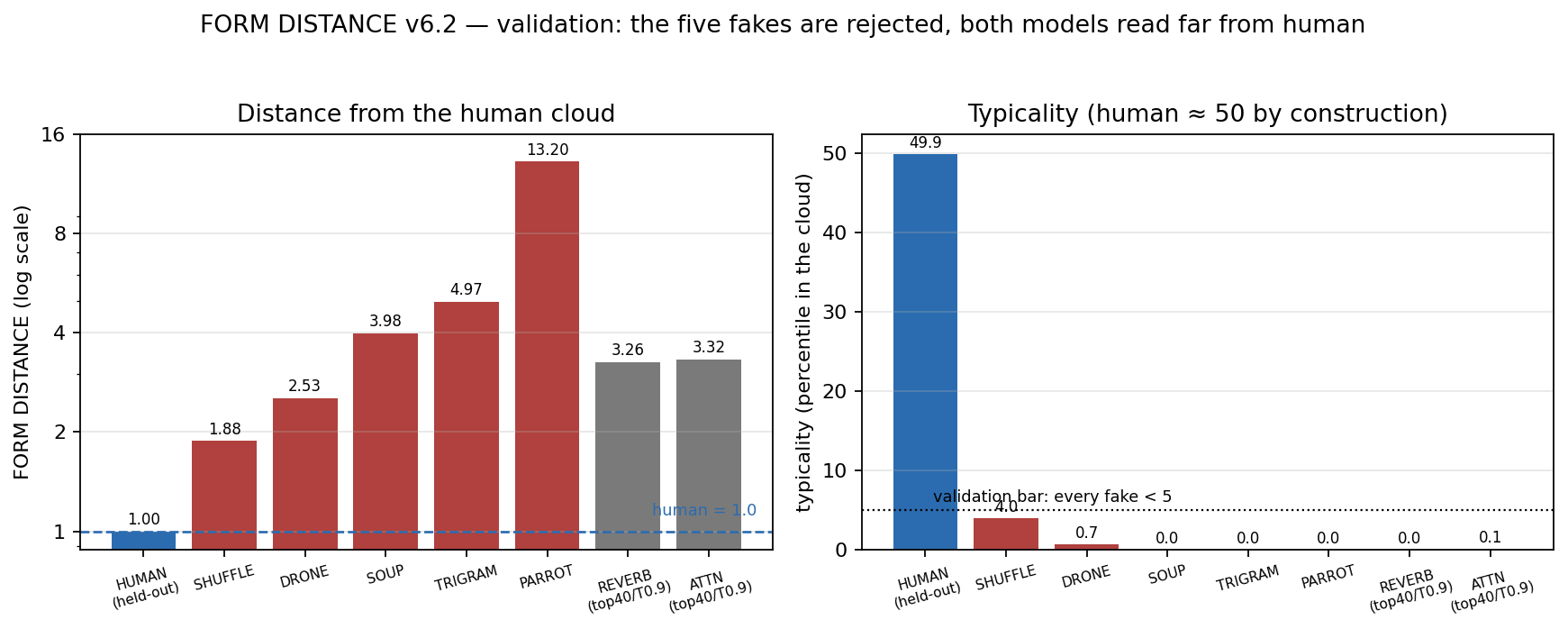}
\caption{Instrument validation: held-out human text vs five
constructed fakes and both conservatively-decoded models, on distance
and typicality.}
\label{fig:meter}
\end{figure}

\paragraph{Validation (Figure~\ref{fig:meter}).} Five constructed
fakes probe the instrument before any model is measured: word soup
(bigram sampling --- locally fluent, no structure), parrot (one
phrase forever), drone (content words overwritten), shuffle (local
word-order destroyed) and trigram babble (corpus-attested triples
stitched into global nonsense --- built specifically to live in the
hole a wide sampler escapes through). All five read far from the
cloud (typicality $\le 4$, most 0.0; distances 1.9--13.2), while
held-out human text reads 1.0/typ${\approx}$50 by construction. The
ninth axis has a history worth telling: an earlier eight-axis version
of the instrument scored a wide-sampler policy as fully human while a
human reader called the same samples word salad; the disagreement
exposed a gaming channel --- invented trigrams --- that CLUNK3 and
the trigram-babble control now close (the full episode, with
formulas, is Appendix~\ref{app:meter}).

\paragraph{First reading.} Both architectures, decoded
conservatively, start at distance ${\approx}$3.3 with the \emph{same}
dominant failure axis (LOCAL: repetition of safe n-grams). The shared
failure is itself a finding: the timidity of small-model text is not
an attention problem, it is a decoding problem --- which is exactly
where Section~\ref{sec:generation} goes next.

\section{Where generation quality actually comes from}
\label{sec:generation}

\subsection{Decoding policy}
\label{sec:decoding}

Twelve-policy sweep (top-$k$ $\times$ temperature) on the frozen
meter: top-40/T0.9 $\to$ 3.17 [2.99, 3.36]; top-1000/T1.15 $\to$
\dist{1.515 [1.451, 1.583]}, typicality 9.1. Same weights ---
decoding alone halves the distance. The conservative standard, not
the model, produced the degenerate repetitive text (LOCAL z: $+5.9
\to -0.1$). The gaming episode belongs here: the eight-axis
instrument read this policy at 1.00 --- statistically human --- while
a human reader called the same samples word salad; the disagreement
exposed the missing axis (CLUNK3, unseen-trigram rate: this policy
sits at $+2.7$\,sd), the instrument was patched and refrozen, and
1.515 is the official baseline (full episode in
Appendix~\ref{app:meter}). What remains is a see-saw no temperature
can fix: policies that cure repetition (LOCAL) pay in invented
trigrams (CLUNK3) --- boldness without three-word coherence.

\begin{figure}[ht]
\centering
\includegraphics[width=0.9\linewidth]{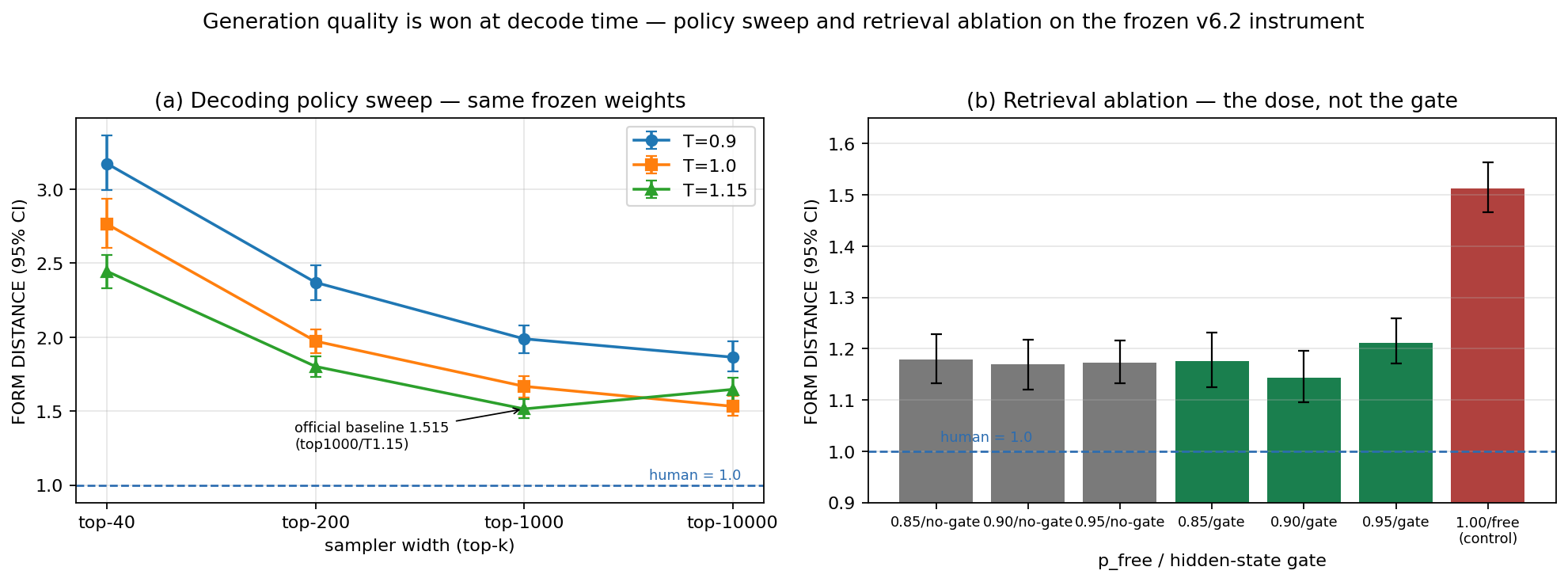}
\caption{Left: the twelve-policy decoding sweep with 95\% CIs.
Right: the retrieval dose $\times$ gate ablation (round 2) against
the free-decoding control.}
\label{fig:sweep}
\end{figure}

\subsection{Retrieval at decode time}
\label{sec:retrieval}

Corpus phrases proposed during generation and interleaved with free
words; a resonance score --- the cosine agreement of the model's own
hidden state over the phrase --- selects among candidates. Clean win,
the first with separated CIs on the frozen instrument: \dist{DIST
1.135 [1.095, 1.175]} vs the 1.515 official baseline, with no
training step involved. The dose-response ablation ($p_{\mathrm{free}}
\in \{0.85, 0.90, 0.95\}$ $\times$ gate on/off) shows the win is a
robust plateau, not a magic point: any sparse phrase dose (phrases
covering 12--32\% of output words) lands at 1.14--1.21, cleanly
separated from the 1.51 control, while gate-on vs gate-off at equal
dose differs within noise. Honest attribution: the gain belongs to
the phrase as a unit --- retrieval itself --- not to the gate;
$p_{\mathrm{free}}=0.90$ with the gate is the best observed operating
point (1.135/1.143 across the two rounds).

\subsection{Memory provenance: the phrases must come from the
model's own corpus}
\label{sec:provenance}

The obvious next question --- does a bigger phrase library help
more? --- has a sharp and initially surprising answer: no; what
matters is not the size of the memory but its provenance. Swapping
the WikiText-2 library for progressively larger slices of
WikiText-103 (2M $\to$ 83M words) makes generation monotonically
\emph{worse} (1.17 $\to$ 1.37 against a 1.49 free control). A
controlled follow-up separates size from source: at matched 2M size
and identical sampling, the model's own corpus gives 1.181 while a
foreign slice gives 1.460 --- statistically indistinguishable from
using no library at all --- and quadrupling the foreign slice does
not help (1.448). Two audits protect the result. First, the
instrument is fair to the foreign domain: held-out \emph{human}
WikiText-103 passages score 1.07 and 0.98 on the same frozen meter,
inside the human cloud, so the deficit belongs to the model, not the
yardstick. Second, the effect is not an artifact of
attention-freedom: the parameter-matched attention twin, trained with
the identical recipe, shows the same gap (own 2M: 1.173; foreign 2M:
1.510; foreign 8M: 1.528; free control 1.618), extracting only a
marginally larger share of value from foreign material (roughly
20--25\% of the own-corpus benefit, versus ${\sim}$10\% for the
recurrent model; the axis signature is shared --- foreign splices
raise the unseen-trigram rate to ${\approx} +1.8$\,sd). We read this
as a scale-of-competence effect rather than an architectural one: the
ability to integrate out-of-distribution material at inference time
(in-context learning, in the large-model literature) is known to
emerge with scale, and at 3M parameters neither architecture has
it --- attention holds at most a small edge. The practical
consequence is stated plainly: at this scale, retrieval is a way to
re-use what the model already knows, not a free upgrade channel for
foreign knowledge; whether the gap closes along the compute-optimal
ladder is an open question we flag in
Section~\ref{sec:conclusion}.

\subsection{What did NOT work (honest negatives, all
pre-registered)}
\label{sec:negatives}

\begin{itemize}
\item TRIO (trigram lesson loss): equivalent to cooling the sampler;
  rejected.
\item HARMOCLIP (reverb-driven harmonic phases): no gain; harmonics
  belong at the byte level.
\item MELODY (FiLM modulation of the output from a raw-embedding
  reverb bus): $+2.7$\% params, no gain (ppl 209 vs 207; DIST 1.53 vs
  1.49). Form is won at decode time, not in the trunk, at this scale.
\item SPELL A/B (computed lexicon: form from bytes via the wavetable
  speller, meaning from counted PPMI vectors, frequency from a
  counted unigram prior): ppl ratio 1.29 vs the learned table (fail
  at our $\le 1.05$ bar) --- but 94\% of the table's top-1 at 19\% of
  its parameters, and it embeds unseen words (nonces) sensibly.
  Buried per protocol; the calibration analysis (rare-word
  probabilities) is the useful residue.
\end{itemize}

\subsection{TALK}

An interactive REPL over the retrieval decoder --- the qualitative
demo that the pieces compose; runs offline on CPU.

\section{Related work}
\label{sec:related}

\begin{itemize}
\item Prior Kathleen work (Papers 1--2 \cite{kathleen2}):
  classification without tokenizer, attention, or pretraining; the
  efficiency-frontier claim carried from there --- ${\sim}$188
  accuracy points per M-params on SST-2, ${\sim}235\times$ fewer
  parameters than BERT-base \cite{devlin2019bert} at comparable task
  accuracy under the full recipe. This paper extends the same
  components to the generation axis.
\item State-space / linear-recurrence models (S4 \cite{gu2022s4},
  Mamba \cite{gu2023mamba}, RWKV \cite{peng2023rwkv}): the reverb
  bank is a content-gated exponential-decay recurrence --- same
  family, arrived at independently from an audio-engineering
  metaphor; our contribution is the byte-level small-scale regime and
  the measurement instrument, not the recurrence itself.
\item Byte-level LMs (ByT5 \cite{xue2022byt5}, MegaByte
  \cite{yu2023megabyte}, the enwik8/Hutter-prize line).
\item Data-scaling methodology follows the small-model-extrapolation
  logic of Kaplan et al.\ \cite{kaplan2020} / Hoffmann et al.\
  \cite{hoffmann2022}, restricted here to the fixed-parameter,
  on-device regime.
\item Text-quality metrics without judges (MAUVE
  \cite{pillutla2021mauve}) --- contrast: FORM DISTANCE is fully
  offline, non-parametric, and validated against constructed fakes.
\item Retrieval-augmented decoding (kNN-LM \cite{khandelwal2020knnlm})
  --- our scheme is the small-scale, phrase-level cousin, and our
  ablation (dose vs gate) is a caution its larger relatives rarely
  run.
\item Holographic Reduced Representations \cite{plate1995hrr} for the
  memory road ahead (Section~\ref{sec:conclusion}).
\end{itemize}

\section{Beyond text: the native-signal argument (evidence, not
claim)}
\label{sec:signals}

The pipeline never tokenizes; it reads signals. For audio this is not
a metaphor: in speaker-independent speech-emotion recognition the
same component family with a gammatone front-end beats a matched
random-conv front-end by \textbf{+11.0pp on RAVDESS
\cite{livingstone2018ravdess} at full data (5 seeds: $44.4\pm1.8$ vs
$33.4\pm2.1$)}, with the advantage present at every data fraction; on
the larger, more natural CREMA-D \cite{cao2014cremad} the gain is
smaller but real and grows with data ($+0.2/+0.8/+0.9$pp at
25/50/100\%). The honest reading: the signal-processing prior is
dataset-dependent but never harmful, and audio is where the
architecture's inductive bias is native. We flag multimodality as the
series' natural next axis (uncompressed pixel streams for vision;
compressed formats such as PNG/JPEG bytes are near-random and out of
scope), and make no omni-model claim here.

\section{Limitations}
\label{sec:limitations}

\begin{itemize}
\item One base corpus for the main road (WikiText-103); enwik8 added
  as robustness, both English.
\item Fixed model size; the flattening curves say capacity, but we
  did not scale parameters with data (compute-optimal frontier is
  future work).
\item Two seeds; small models are noisy, CIs reported where we have
  them.
\item The attention baseline is our own parameter-matched skeleton,
  not a tuned external transformer.
\item FORM DISTANCE measures form, not truth or coherence of content.
\item Retrieval helps only from the model's own training corpus at
  this scale (Section~\ref{sec:provenance}); we did not test whether
  the own-vs-foreign gap closes as parameters and data grow together.
\end{itemize}

\section{Conclusion and the road ahead}
\label{sec:conclusion}

Same ingredients, new verb: the attention-free byte pipeline of
Papers 1--2 \emph{writes}. In the regime it was designed for ---
sub-million parameters, megabytes-to-gigabytes of data, free-tier or
on-device hardware, offline --- it reads text better than a
parameter-matched transformer at every scale measured, and decoding
policy plus retrieved phrases close most of the form gap ($3.2 \to
1.14$ on a validated, gaming-patched instrument) without a single
training step. The recurrence also inherits the family's serving
economics: $O(L)$ time and a constant-size state instead of a
KV-cache that grows with context --- an advantage that compounds at
byte level, where sequences are several times longer than their
tokenized equivalents.

Three open questions, in order: \textbf{(1) Compute-optimal
scaling} --- both curves here flatten toward the same fixed-capacity
ceiling; the decisive experiment grows parameters and data together
(and only there does testing oscillator coupling across banks become
meaningful); the same ladder carries a second question, whether the
own-vs-foreign retrieval gap of Section~\ref{sec:provenance} closes
as competence grows --- if it does, the ladder measures the emergence
of memory generality. \textbf{(2) Explicit memory} --- attention is a
lossless lookup table over the whole context, and that is precisely
what a fading recurrent state, a lossy compressor by design, cannot
be: exact long-range recall is the known weakness of this model
family. Our answer is not to reintroduce attention but to pair the
stream with the complementary mechanism: holographic (HRR) key-value
binding \cite{plate1995hrr} with learned write/read gates, which
already passes isolated recall proofs (0.95+ cosine recall from
100K-position haystacks at $O(1)$ memory) and awaits integration into
the trunk --- the fading stream for form, the hologram for facts;
Section~\ref{sec:provenance} adds a design requirement, that the
written memory must remain usable even when its content is
out-of-distribution for the trunk. \textbf{(3) Native signals beyond
text} --- Section~\ref{sec:signals}. The series continues: reads,
reads without a teacher, writes --- next, remembers and hears.

\appendix

\section{Fast reverb scan}
\label{app:scan}

The original ReverbBank walks the recurrence one step at a time in a
Python loop (256 kernel launches per 256-byte window). We replace it
with a Mamba-style chunked closed form: within a chunk of $C$ steps,
\begin{align*}
s_t &= c_t \cdot s_0 + c_t \sum_{i \le t} \frac{(1-g_i)\,v_i}{c_i},
& c_t &= \prod_{i \le t} g_i ,
\end{align*}
which is a cumprod plus a cumsum, fully vectorised; a single Python
step per chunk carries the state across chunk boundaries. With $C=16$
and $\gamma \ge 0.5$ the ratio $1/c_i \le 2^{16}$, safely inside
float32 (the production Kaggle build uses $C=16$; the benchmark build
used $C=64$).

\textbf{Equivalence} (same weights, state dicts interchangeable): max
absolute difference vs the sequential loop $8.6 \times 10^{-7}$
(relative $8.0 \times 10^{-7}$); through the MultiScaleReverb wrapper
$2.4 \times 10^{-7}$; the padded path ($L$ not divisible by the
chunk) tested separately. \textbf{Throughput:} $\times 25.4$ over the
loop at $L = 16{,}384$ (10--50$\times$ across the lengths measured).
Honest note: the scan is still $\times 2.7$ behind the attention
baseline in wall-clock at that length --- closing that gap needs a
fused kernel, and is orthogonal to every claim in this paper
(bits-per-byte is hardware-independent). What the scan buys here is
concrete: the 512\,MB training point completes inside a single free
Kaggle T4 session.

\section{FORM DISTANCE axes, formulas, and the gaming episode}
\label{app:meter}

Axis definitions (verbatim from the frozen v6.2 instrument; sp = the
corpus statistics space, content = non-glue words):

\begin{itemize}
\item RICH: distinct content words / total content words.
\item FLOW: distinct word-trigrams / total word-trigrams.
\item HOLD: cosine between the mean vectors of the opening and
  closing thirds of the passage.
\item GRAIN: mean cosine between adjacent 10-word chunks.
\item NEW: share of closing-third content words unseen earlier in the
  passage.
\item LOCAL: mean log-probability of adjacent word pairs under corpus
  bigram statistics.
\item ARC: slope of semantic distance between chunks against their
  separation (development: real writing drifts; a parrot and a soup
  are both flat).
\item CLUNK: share of adjacent pairs never seen in the corpus.
\item CLUNK3: share of word-triples never seen in the corpus
  (two-sided, like CLUNK: trigram babble reads ${\sim}$0, salad reads
  high).
\end{itemize}

No axis has a ``good'' direction: an axis is kept only if both
degenerate controls (soup, parrot) fall outside the human band on it,
in opposite directions where applicable. The score is the distance to
the 400-passage human cloud under the axis covariance, normalized so
held-out human text reads 1.0; if a candidate ever outscores the
human reference, the instrument prints an alarm instead of a number.

\textbf{The gaming episode, in full.} On the eight-axis instrument
the policy top-1000/T1.15 read 1.002 [0.941, 1.060] ---
statistically human. A human reader called the same samples word
salad. Diagnosis: the policy had cured repetition (LOCAL z: $+5.9 \to
-0.1$) by inventing trigrams --- every pair attested, the triples
nonsense --- a channel no existing axis measured. The patch was
two-sided: a ninth axis (CLUNK3) and a sixth validation control
(trigram babble: corpus-attested triples stitched into global
nonsense, built to live exactly in that hole; it must read $< 5$
typicality or the instrument fails its own validation). On the
refrozen v6.2 instrument the same policy reads 1.515 [1.451, 1.583]
with CLUNK3 at $+2.7$\,sd, and the trigram control reads 0.0. The
lesson we draw is methodological: a fixed instrument plus an
attentive human is a gaming-detection loop; each disagreement becomes
an axis, and the instrument is refrozen before any further
measurement.

\section{Full experiment registry}
\label{app:registry}

Every generation-axis experiment was pre-registered (verdict criteria
written down before running) and is preserved with its script and
JSON report:

\begin{table}[ht]
\centering
\small
\caption{Experiment registry.}
\label{tab:registry}
\begin{tabular}{p{0.30\linewidth}p{0.34\linewidth}p{0.28\linewidth}}
\toprule
Experiment & Verdict & Key numbers \\
\midrule
Decoding sweep (12 policies) & ADOPTED: top-1000/T1.15 & $3.17 \to 1.515$ \\
Instrument validation & PASS (all 5 fakes rejected) & typ $\le 4$; human 49.9 \\
LICKS round 1 (retrieval) & CLEAN WIN & 1.135 [1.095, 1.175] vs 1.513 \\
LICKS round 2 (dose $\times$ gate) & WIN = dose; gate within noise & plateau 1.14--1.21 \\
LICKS@103 round A (library scale) & MISMATCH: bigger foreign library hurts & 2M 1.17 $\to$ 83M 1.37 (ctrl 1.49) \\
LICKS@103 round B (size vs source) & SOURCE: provenance, not size & own-2M 1.181; foreign-2M 1.460 $\approx$ ctrl \\
Meter domain audit & METER IS FAIR to foreign human text & human WT-103: 1.07 / 0.98 \\
LICKS / attention twin & SAME GAP; marginal attention edge & own 1.173; foreign 1.510/1.528 (ctrl 1.618) \\
TRIO (trigram lesson loss) & REJECTED ($\equiv$ cooling the sampler) & --- \\
HARMOCLIP (harmonic phases) & REJECTED (no gain) & --- \\
MELODY (FiLM reverb bus) & REJECTED ($+2.7$\% params, no gain) & ppl 209 vs 207 \\
SPELL A/B (computed lexicon) & FAIL at the $\le 1.05$ bar & ratio 1.29; 94\% top-1 at 19\% params \\
Fast reverb scan & PASS (equivalence + speed) & $8.6\mathrm{e}{-7}$; $\times 25.4$ \\
Scaling road (5 points) & REVERB ahead everywhere & Table~\ref{tab:road} \\
Robustness (seed 43, enwik8) & REPRODUCES & Tables~\ref{tab:seeds}--\ref{tab:enwik8} \\
BRIDGE (frozen probe) & THE CHAIN HOLDS & Table~\ref{tab:bridge} \\
\bottomrule
\end{tabular}
\end{table}


\begin{thebibliography}{99}

\bibitem{devlin2019bert}
J.~Devlin, M.-W.~Chang, K.~Lee, K.~Toutanova.
\newblock BERT: Pre-training of Deep Bidirectional Transformers for
Language Understanding.
\newblock In \emph{NAACL}, 2019.

\bibitem{gu2022s4}
A.~Gu, K.~Goel, C.~R\'e.
\newblock Efficiently Modeling Long Sequences with Structured State
Spaces.
\newblock In \emph{ICLR}, 2022.

\bibitem{gu2023mamba}
A.~Gu, T.~Dao.
\newblock Mamba: Linear-Time Sequence Modeling with Selective State
Spaces.
\newblock arXiv:2312.00752, 2023.

\bibitem{peng2023rwkv}
B.~Peng et al.
\newblock RWKV: Reinventing RNNs for the Transformer Era.
\newblock In \emph{Findings of EMNLP}, 2023.

\bibitem{xue2022byt5}
L.~Xue et al.
\newblock ByT5: Towards a Token-Free Future with Pre-trained
Byte-to-Byte Models.
\newblock \emph{TACL}, 2022.

\bibitem{yu2023megabyte}
L.~Yu et al.
\newblock MegaByte: Predicting Million-Byte Sequences with Multiscale
Transformers.
\newblock In \emph{NeurIPS}, 2023.

\bibitem{kaplan2020}
J.~Kaplan et al.
\newblock Scaling Laws for Neural Language Models.
\newblock arXiv:2001.08361, 2020.

\bibitem{hoffmann2022}
J.~Hoffmann et al.
\newblock Training Compute-Optimal Large Language Models.
\newblock In \emph{NeurIPS}, 2022.

\bibitem{pillutla2021mauve}
K.~Pillutla et al.
\newblock MAUVE: Measuring the Gap Between Neural Text and Human Text
using Divergence Frontiers.
\newblock In \emph{NeurIPS}, 2021.

\bibitem{khandelwal2020knnlm}
U.~Khandelwal, O.~Levy, D.~Jurafsky, L.~Zettlemoyer, M.~Lewis.
\newblock Generalization through Memorization: Nearest Neighbor
Language Models.
\newblock In \emph{ICLR}, 2020.

\bibitem{plate1995hrr}
T.~A.~Plate.
\newblock Holographic Reduced Representations.
\newblock \emph{IEEE Transactions on Neural Networks}, 1995.

\bibitem{merity2017wikitext}
S.~Merity, C.~Xiong, J.~Bradbury, R.~Socher.
\newblock Pointer Sentinel Mixture Models.
\newblock In \emph{ICLR}, 2017.

\bibitem{socher2013sst}
R.~Socher et al.
\newblock Recursive Deep Models for Semantic Compositionality over a
Sentiment Treebank.
\newblock In \emph{EMNLP}, 2013.

\bibitem{maas2011imdb}
A.~L.~Maas et al.
\newblock Learning Word Vectors for Sentiment Analysis.
\newblock In \emph{ACL}, 2011.

\bibitem{livingstone2018ravdess}
S.~R.~Livingstone, F.~A.~Russo.
\newblock The Ryerson Audio-Visual Database of Emotional Speech and
Song (RAVDESS).
\newblock \emph{PLOS ONE}, 2018.

\bibitem{cao2014cremad}
H.~Cao et al.
\newblock CREMA-D: Crowd-Sourced Emotional Multimodal Actors Dataset.
\newblock \emph{IEEE Transactions on Affective Computing}, 2014.

\bibitem{kathleen2}
G.~Fountzoulas.
\newblock Kathleen: Oscillator-Based Byte-Level Text Classification
Without Tokenization or Attention.
\newblock arXiv:2604.07969, 2026.

\end{thebibliography}
\end{document}